\documentclass[10pt,a4paper]{article}

\usepackage[T1]{fontenc}
\usepackage[utf8]{inputenc}
\usepackage{amsmath,amssymb,amsfonts}
\usepackage{graphicx}
\usepackage[export]{adjustbox}
\usepackage{geometry}
\usepackage{times}
\usepackage[numbers,sort&compress]{natbib}
\usepackage{caption}
\usepackage{enumitem}
\usepackage{float}
\usepackage{url}
\usepackage{booktabs}
\usepackage{hyperref}

\hypersetup{hidelinks}

\def\keywords#1{\par\noindent\textbf{Keywords:} #1\par\vskip 0.5cm}
\newcommand{\bioentry}[4]{%
\noindent
\begin{minipage}[t]{0.15\linewidth}
\centering
\includegraphics[height=3.6cm,keepaspectratio,valign=t]{#1}
\end{minipage}\hfill
\begin{minipage}[t]{0.78\linewidth}
\raggedright
\textbf{#2} #3\\
\textit{E-mail: }\nolinkurl{#4}
\end{minipage}\par\vspace{1.5em}
}

\begin{document}

\begin{center}
{\LARGE\bfseries AIGC video detection based on the fusion of spatial-frequency-optical flow multimodal features\par}
\vspace{1em}
{\large HONG Sheng$^{1,*}$, WANG Xuanqi$^2$, ZHANG Chang$^3$, WANG Jiacheng$^1$, DUAN Pingxia$^4$, and WANG Yuwei$^5$\par}
\vspace{0.8em}
{\small
$^1$School of Cyber Science and Technology, Beihang University, Beijing 100191, China\\
$^2$School of Information and Engineering, Nanchang University, Nanchang 330031, China\\
$^3$Continuing Education College, Beihang University, Beijing 100191, China\\
$^4$DBAPPSecurity Co., Ltd. (DAS-Security), Beijing 100195, China\\
$^5$Institute of Computing Technology, Chinese Academy of Sciences, Beijing 100190, China
}
\end{center}

\begin{abstract}
The rapid evolution of generative artificial intelligence (AI) (e.g., Sora, Hunyuan) makes it essential to develop effective detection strategies that can generalize across ever-evolving synthesis techniques. This study is motivated by the observation of a fundamental challenge in generative models: the inherent difficulty of maintaining cross-modal consistency between appearance and motion. To this end, we propose a multi-modal framework for artificial intelligence generated content (AIGC) video forgery detection, named cross-attention based video forgery detector (CrossAtt-VFD), based on joint multi-view analysis of content. Methodologically, we introduce a dual-branch architecture that simultaneously extracts spatial-frequency and optical-flow features. This approach enables the modeling of videos from complementary perceptual perspectives. The core of this process is a dedicated cross-attention mechanism, which governs the alignment of the two modalities and translates cross-modal inconsistencies into a potent diagnostic signal. This multi-modal strategy facilitates the detection of motion that is statistically inconsistent with the visual appearance of a scene. Comprehensive experimental results demonstrate that our model achieves an accuracy of 94.32\%, a precision of 91.67\%, and a recall of 96.25\%, effectively verifying the advantages of the multi-modal fusion strategy.
\end{abstract}


\noindent Manuscript received January 09, 2026.\\
\noindent *Corresponding author.\\
\noindent This work was supported by the National Key Research and Development Program of China (2022YFB3103602).\\
\noindent DOI: 10.23919/JSEE.2026.000049

\section{Introduction}
The emergence of artificial intelligence generated content (AIGC) represents a paradigm shift in visual content production. While it enables unprecedented modes of expression, it is ultimately a double-edged sword. The rise of industry-leading models such as Sora, Kling, and Hunyuan has democratized the mass production of high-fidelity media \cite{rf1,rf2}. This accessibility, however, also facilitates its reckless misuse, enabling the widespread deception of the public. With the continuous advancement of generative technologies, this paper incorporates the abuse of AIGC-synthesized fake content into the broader scope of video forgery. Therefore, developing effective detection mechanisms for AIGC content has become crucial to protect public trust and maintain the integrity of information in the digital world.

Building on this demand for effective detection, recent deep learning techniques have primarily focused on learning discriminative spatiotemporal representations from videos. Progress in this field is reflected in several key developments. The incorporation of domain generalization and auxiliary supervision signals has significantly enhanced the robustness of deepfake detection models \cite{rf3}. Meanwhile, the rapid evolution of generative techniques themselves, including those capable of producing content in complex and open-world scenarios, continuously raises the bar for detection methods. Foundational forensic principles, such as analyzing feature consistency across different granularities \cite{rf5}, remain highly relevant. The security and robustness of deep features are equally paramount, prompting research into how convolutional neural network (CNN) architectures respond to adversarial threats \cite{rf6}. The field is further systematized by comprehensive reviews that summarize the landscape of AI-created visual content and corresponding detection technologies \cite{rf7}. In addition, novel detection frameworks have been proposed to handle emerging challenges like multi-definition deepfakes via semantics reduction and cross-domain training \cite{rf8}. These advancements collectively shape the methodological discourse and form the foundation of our work. More recently, the community has begun to emphasize interpretability and broader multimodal benchmarks. For instance, IVY-FAKE introduces a unified explainable framework that enhances the transparency and credibility of detection results by providing detailed annotations on why a sample is judged as real or fake \cite{rf9}. Similarly, recent works focusing on missing motion details align with our findings on motion anomalies \cite{rf10}, while comprehensive datasets highlight the growing trend of incorporating audio-visual multimodal cues for AIGC detection \cite{rf11}. Our objective is to effectively embed cross-modal inconsistencies, representing a more efficient detection mechanism.

Nevertheless, while significant progress has been made with such approaches, they typically rely on a single or limited modal perspective (e.g., spatial or temporal). This makes them vulnerable to sophisticated generators that successfully replicate artifacts in one domain but fail in another. In this paper, we propose a more robust and comprehensive approach based on learned multi-modality, which successfully integrates and exploits mutually complementary perspectives from the video data itself. By simultaneously analyzing the spatial domain, frequency domain, and motion dynamics, and by critically examining the correlation of responses across these modalities, our method obtains a detector that captures and reveals deeper, more generalizable representations of synthesis quality---representations that remain hidden when each viewpoint is considered in isolation.

Given these risks, establishing effective detection mechanisms for AIGC is critical to mitigating societal harms and upholding information authenticity. The main contributions of this work are as follows:
\begin{enumerate}[label=(\roman*)]
\item We propose and validate that ``cross-modal inconsistency'' is a more generalizable and robust footprint for detecting AIGC videos than artifacts in any single modality.
\item We design a multi-modal fusion framework. Its core is a cross-attention module that acts not merely as a fusion tool but as an ``inconsistency diagnostic engine'' dynamically modeling the asymmetric relationship between appearance (spatial-frequency) and motion (optical flow) features.
\item We construct a comprehensive dataset that includes videos generated by multiple state-of-the-art (SOTA) models. Experiments on this dataset show that our framework achieves an accuracy of 94.32\%, a precision of 91.67\%, and a recall of 96.25\%. These results substantiate its superior performance and enhanced generalization capability, particularly against unseen generative models.
\end{enumerate}

\section{Relevant theoretical and technical foundations}
\subsection{Common flaws in generative videos}
Though traditional video forensics approaches provide a basis for understanding media integrity, they are unable to address the specific issues posed by AIGC \cite{rf12}. Thus, our work rests on the premise that the shortcomings of contemporary video generation models manifest as common artifacts in three specific areas. It is the very predictability of these shortcomings that enables the construction of a highly generalizable detection model.

\subsubsection{Violation of physical laws}
Although generative models like Sora can produce highly realistic visual content, this capability stems primarily from statistical pattern association rather than a genuine understanding of physical dynamics \cite{rf13}. This inherent limitation manifests during the generation process itself, often resulting in content that violates fundamental physical principles. These violations can be categorized as follows:
\begin{enumerate}[label=(\roman*)]
\item Non-coherent motion: an object may exhibit sudden, unnatural accelerations or decelerations, or its direction of motion may change erratically across frames.
\item Geometric inconsistencies: these often arise from violations of perspective principles or inconsistent occlusion relationships between consecutive frames. A common cause is the lack of accurate depth information, leading to implausible occlusion events---for instance, a foreground object being incorrectly occluded by a background element. Such artifacts can be effectively identified through panoptic-aware depth estimation methods \cite{rf14}.
\item Violations of physics: the generated content depicts events that violate fundamental laws of physics.
\end{enumerate}

\subsubsection{Defects in temporal coherence}
\begin{figure}[H]
\centering
\includegraphics[width=0.62\linewidth]{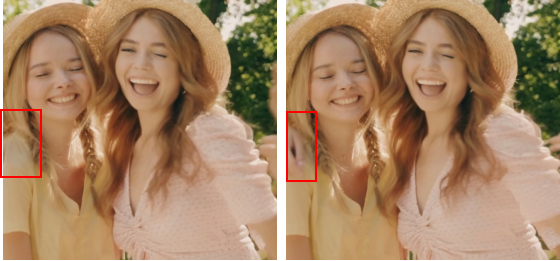}
\caption{Example of inter-frame mutation in AIGC video. (a) Normal scene of the hand not yet appearing. (b) Sudden hand deformity appearing on the shoulder area.}
\label{fig:interframe_mutation}
\end{figure}

Defective modeling of long-range dependencies typically results in temporal inconsistencies, which manifest primarily in two forms:
\begin{enumerate}[label=(\roman*)]
\item Inter-frame mutations: objects may exhibit non-smooth motion, or scenes may transition in an abrupt and unnatural manner. As shown in Fig.~\ref{fig:interframe_mutation}, in an AIGC video case, the originally absent hand suddenly appears on another person's shoulder within an extremely short frame interval. This instantaneous, physically unfounded mutation reveals the model's deficiency in handling cross-frame consistency.
\item Missing long-range dependencies: longer video sequences may exhibit logical discontinuities. For example, the scene may illogically jump from an indoor to an outdoor setting without any coherent transition.
\end{enumerate}

\subsubsection{Abnormal frequency-domain characteristics}
The diffusion process, fundamental to many existing generative models, introduces specific artifacts that are particularly evident in the frequency domain, such as the loss of high-frequency details. To capture these discriminative patterns, recent research has developed networks that explicitly leverage frequency-domain transformations. For instance, sequency discrete cosine transform convolution networks have been designed to enhance visual recognition by intrinsically incorporating frequency analysis \cite{rf15}. Consequently, the frequency-domain deficiencies inherent to generated content can be effectively identified by such specialized architectures.

Specific manifestations include the following:
\begin{enumerate}[label=(\roman*)]
\item Loss of high-frequency detail: the attenuation of high-frequency components results in blurred edges, diminished textural details, and an unnaturally smooth appearance.
\item Unnatural color banding: the degradation of subtle color transitions can cause clearly defined banding where smooth gradients should exist.
\end{enumerate}
Collectively, these three categories of defects form a distinctive fingerprint for AIGC videos and provide the theoretical foundation for the multi-modal detection framework proposed in this study.

\section{Methodology}
In this section, motivated by the analysis of common generative flaws in Section 2, we propose a framework for video detection that leverages multimodal spatiotemporal features. The core premise of our approach is that the integration of complementary multimodal information---including spatial appearance, frequency-domain signatures, and motion dynamics (optical flow)---enables more effective detection. Our framework enables the identification of a broader spectrum of artifacts, ranging from static texture anomalies to dynamic physical implausibilities. This facilitates the learning of a more comprehensive and generalizable representation for detecting AIGC videos.

\subsection{Multi-modal framework overview}
In this section, we propose cross-attention based video forgery detector (CrossAtt-VFD), a multi-modal video forgery detection framework that leverages cross-attention to mine spatiotemporal inconsistencies. To efficiently investigate the inherent differences between AIGC and real videos, we propose a multimodal detection framework based on three synergistic modalities: spatiotemporal appearance, frequency-domain signatures, and optical flow.

\begin{figure}[H]
\centering
\includegraphics[width=\linewidth]{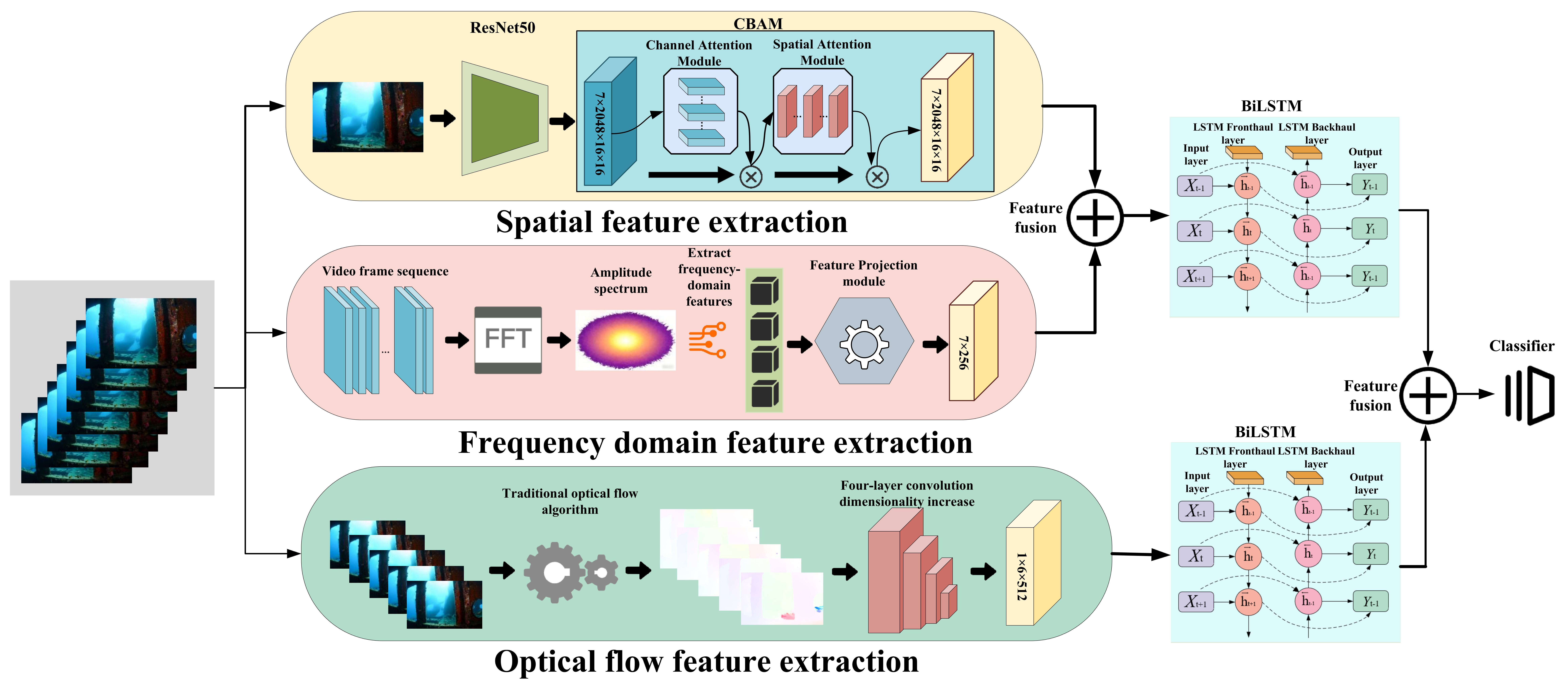}
\caption{Multi-modal framework overview.}
\label{fig:framework}
\end{figure}

As detailed previously, our detection framework employs a two-stream architecture, illustrated in Fig.~\ref{fig:framework}, which outlines the parallel architecture of the spatial-frequency and optical flow branches. This design is inspired by the classic two-stream network \cite{rf16}. The spatial-frequency stream uses a CNN to extract static features from each video frame. These features are subsequently transformed into a frequency-domain representation. A bidirectional long short term memory (BiLSTM) network then models the temporal evolution and consistency of these combined spatial-frequency features. The optical flow stream computes sparse optical flow between consecutive frames. A separate BiLSTM then encodes these motion trajectories to model temporal dynamics, paralleling the process in the spatial-frequency stream. All input videos are uniformly sampled to a fixed number of frames and resized to a $512\times512$ pixel resolution. The frame sequences are then normalized.

Our framework employs a proven two-branch architecture, as shown in Fig.~\ref{fig:framework}, which is particularly effective for processing heterogeneous data streams in parallel within multimodal learning. This design is essential for our multi-view approach, enabling one branch to specialize in spatial appearance while the other focuses exclusively on motion dynamics, thereby ensuring dedicated domain-specific analysis. The necessity of specialized branches is further supported by recent trends in video super-resolution, such as MambaOVSR, which utilizes multiscale fusion and global motion modeling to preserve temporal integrity \cite{rf34}. Our framework adopts a similar multi-view philosophy to ensure dedicated analysis of motion dynamics versus spatial appearance. At the heart of our multimodal fusion strategy lies a dedicated cross-attention mechanism. This module functions not as a mere fusion tool, but as an ``inconsistency diagnostic engine'' that dynamically reasons about the relationships between different views. Specifically, it allows the spatial-frequency stream (the ``appearance view'') to query the optical flow stream (the ``motion view''), thereby identifying regions where the observed motion is physically implausible given the visual context.

The two streams are finally fused using the aforementioned attention mechanism, which has been successfully applied in related tasks such as detecting compression artifacts in fake videos \cite{rf17}. It dynamically integrates the two feature types to produce a final detection score. We denote the temporal feature vector from the spatial-frequency branch (via BiLSTM) as $F_{\mathrm{sf}}$, and the motion feature vector from the optical flow branch as $F_{\mathrm{of}}$. The fusion of these two vectors is derived as follows:
\begin{equation}
A_{\mathrm{of}\rightarrow\mathrm{sf}}=\mathrm{Softmax}\left(\frac{F_{\mathrm{sf}}F_{\mathrm{of}}^{T}}{\sqrt{d_k}}\right)
\end{equation}
where $d_k$ is the dimensionality of the key vectors, and $A_{\mathrm{of}\rightarrow\mathrm{sf}}$ represents the importance weight of different dimensions in the optical flow features relative to the spatial-frequency features. These weights are then used to compute the weighted aggregation of the optical flow features, yielding optical-flow attention features that are adapted to the spatial-frequency characteristics:
\begin{equation}
F_{\mathrm{of}}^{\mathrm{att}}=A_{\mathrm{of}\rightarrow\mathrm{sf}}F_{\mathrm{of}}.
\end{equation}
The final fused features are then passed through the classification head to produce the output prediction:
\begin{equation}
S_{\mathrm{final}}=\mathrm{FC}\left(F_{\mathrm{of}}^{\mathrm{att}}\right).
\end{equation}

\subsection{Spatial feature extraction module}
Our framework leverages a pre-trained ResNet to extract rich spatial features from video frames, which accelerates convergence \cite{rf18}. These features are then fed into a BiLSTM to model temporal dependencies. Thanks to its bidirectional structure, the BiLSTM captures contextual information from both past and future frames, thereby effectively learning spatiotemporal relationships and identifying subtle temporal incoherencies in video sequences.

To meet the objective of developing a lightweight detection model deployable on standard office equipment, and considering hardware constraints, we adopt ResNet-50 as the spatial feature extractor. Our preliminary experiments compared ResNet-18 and ResNet-50 as backbones. The results indicated that while ResNet-50 incurs a slight increase in memory usage, it provides superior convergence speed and generalization ability, achieving higher accuracy and F1-scores compared to the shallower ResNet-18. Therefore, ResNet-50 offers an optimal balance between feature extraction depth and computational efficiency. The original classification head (comprising global pooling and fully connected layers) is removed, while the backbone structure from conv1 to layer4 is retained. This outputs 2\,048-dimensional spatial features. The use of multi-level convolutional features serves this objective, as it has been demonstrated that a multi-scale feature hierarchy significantly enhances the model's capability to capture both subtle artifacts (e.g., blurred edges) and higher-level semantic inconsistencies.

For a given input video clip $V\in\mathbb{R}^{T\times H\times W\times 3}$, where $T$ is the number of frames and $H\times W$ is the spatial resolution, features are first extracted independently from each frame:
\begin{equation}
f_t=\mathrm{ResNet50}_{\mathrm{trunc}}(I_t)
\end{equation}
where $\mathrm{ResNet50}_{\mathrm{trunc}}$ represents the modified network with its final classification head and pooling layers removed, $I_t$ is the $t$-th frame image, and $f_t$ is the extracted spatial feature corresponding to the frame. These spatial features are then stacked along the temporal dimension to form the overall spatial feature representation of the video clip:
\begin{equation}
F=\mathrm{Stack}([f_1,f_2,\ldots,f_T]).
\end{equation}

While the extracted spatial features are rich in visual information, they also contain redundant noise. Furthermore, subtle artifacts indicative of video generation are often obscured within these complex feature maps. To fully exploit the discriminative information within these features, our module implements a two-stage strategy: first, the convolutional block attention module (CBAM) enhances salient visual patterns and suppresses noise. While we employ CBAM for pattern enhancement, recent architectures such as PH-Mamba have further optimized feature representation by harmonizing attention mechanisms with precise position encoding \cite{rf33}. This underscores the importance of structured attention in capturing the subtle artifacts obscured within complex feature maps. Subsequently, a BiLSTM captures the temporal dynamics and long-range dependencies within the refined feature sequence.

\subsection{Frequency-domain feature extraction module}
Modern video generation models, including Sora, Hunyuan, and Wanxiang, are predominantly built on diffusion principles. Their inherent multi-step iterative denoising process introduces distinct artifacts in the frequency domain of synthesized videos when compared to real footage. To capture these artifacts, we design a feature extraction module based on the fast Fourier transform (FFT). The output of this module is subsequently fused with the spatial features to collectively enhance the overall performance of video forensic detection. Recent advancements in image restoration, such as FMRNet, have demonstrated the power of frequency mutual revision to iteratively refine features across different spectral domains \cite{rf32}. Similarly, our module decomposes video frames to identify spectral anomalies characteristic of AIGC.

The module first decomposes video frames into the frequency domain via the FFT. It then employs convolutional feature learning to identify subtle anomalies in the spectral distribution of AIGC videos, specifically targeting the characteristic high-frequency detail loss, to generate highly discriminative frequency-domain feature vectors. The input to this module is a video frame sequence $X$ with dimensions $[B,T,C,H,W]$, where $B$ is the batch size, $T$ is the sequence length, $C$ is the number of channels, and $H$ and $W$ are the frame height and width, respectively. The following operations are performed to convert spatial information into the frequency domain:
\begin{enumerate}[label=(\roman*)]
\item Dimension reshaping: the video frame sequence is reshaped into a batch of individual frames with shape $[B\times T,C,H,W]$, to facilitate independent frequency-domain analysis for each frame.
\item Fourier transform: the FFT is applied to each individual frame. The image is mapped from the spatial domain to the frequency domain, yielding a complex-valued spectrum tensor:
\begin{equation}
\hat{x}=\mathrm{fft2}(x_{\mathrm{frame}}).
\end{equation}
To facilitate the analysis of low-frequency components (corresponding to smooth regions in the image) and high-frequency components (corresponding to edges and details), the obtained spectrum is shifted using \texttt{fftshift} to relocate the low-frequency components to the center of the spectrum, yielding $\hat{x}_{\mathrm{shift}}$. The physical significance of the spectrum is primarily represented by its magnitude spectrum:
\begin{equation}
A=\left|\hat{x}_{\mathrm{shift}}\right|.
\end{equation}
Here, $A$ is a real-valued magnitude spectrum tensor with shape $[B\times T,1,128,128]$, where each element represents the energy intensity of the corresponding frequency component.
\item Frequency-domain statistic extraction: based on the magnitude spectrum $A$, the module extracts four categories of frequency-domain statistical features to determine whether a video is AIGC. These are total energy, high-frequency energy ratio, frequency-domain entropy, and central low-frequency ratio.
\begin{enumerate}[label=\alph*.]
\item Total energy: this metric measures the overall energy intensity of the spectrum. It is mathematically defined as the sum of all elements in the magnitude spectrum:
\begin{equation}
E_{\mathrm{total}}=\sum_{i=1}^{128}\sum_{j=1}^{128}A[i,j]+\epsilon
\end{equation}
where $\epsilon=10^{-8}$ is a small constant to avoid division by zero.
\item High-frequency energy ratio: this metric reflects the relative richness of high-frequency components. Taking the spectrum center $(h_c,w_c)$ as the origin, the Euclidean distance from each frequency point $(i,j)$ to the center is calculated as
\begin{equation}
d(i,j)=\sqrt{(i-h_c)^2+(j-w_c)^2}.
\end{equation}
Define the high-frequency mask $M_{\mathrm{high}}$ as the region where the distance is greater than the threshold:
\begin{equation}
M_{\mathrm{high}}[i,j]=
\begin{cases}
1, & d(i,j)>d_{\max}\cdot \mathrm{high\_freq\_threshold}\\
0, & \text{otherwise}.
\end{cases}
\end{equation}
The high-frequency energy ratio is then
\begin{equation}
r_{\mathrm{high}}=\frac{\sum_{i,j}A[i,j]\cdot M_{\mathrm{high}}[i,j]}{E_{\mathrm{total}}}.
\end{equation}
\item Frequency-domain entropy: this metric measures the randomness of the spectral energy distribution. First, the amplitude spectrum is normalized into a probability distribution
\begin{equation}
P[i,j]=\frac{A[i,j]}{E_{\mathrm{total}}}.
\end{equation}
The frequency-domain entropy is then defined as
\begin{equation}
H=-\sum_{i,j}P[i,j]\log(P[i,j]+\epsilon).
\end{equation}
\item Central low-frequency ratio: this metric reflects the concentration of low-frequency components. The central mask $M_{\mathrm{center}}$ is defined as
\begin{equation}
M_{\mathrm{center}}[i,j]=
\begin{cases}
1, & d(i,j)<0.4d_{\max}\\
0, & \text{otherwise}.
\end{cases}
\end{equation}
The central low-frequency ratio is then
\begin{equation}
r_{\mathrm{center}}=\frac{\sum_{i,j}A[i,j]\cdot M_{\mathrm{center}}[i,j]}{E_{\mathrm{total}}}.
\end{equation}
\end{enumerate}
\item Feature enhancement and output: the processed frequency-domain features are stacked into a feature tensor of shape $[B\times T,4]$. This tensor is then enhanced using a feature projection module, ultimately outputting a frequency-domain feature vector with dimensions $[B\times T,128]$.
\end{enumerate}

\subsection{Optical flow feature extraction module}
The motion in AIGC videos is often unnatural and frequently violates physical laws. Phenomena such as object suspension or instantaneous, inertia-free state transitions can only be detected by analyzing optical flow, which serves as a crucial indicator for these artifacts. In contrast to computationally heavy, data-demanding learning-based methods, we employ classical optical flow algorithms. These methods are computationally efficient, do not require annotated data for training, and are supported by a solid theoretical foundation regarding their convergence and reliability \cite{rf20}. This approach is well-suited to our hardware constraints while providing a robust estimate of motion inconsistencies.

\begin{figure}[H]
\centering
\includegraphics[width=0.9\linewidth]{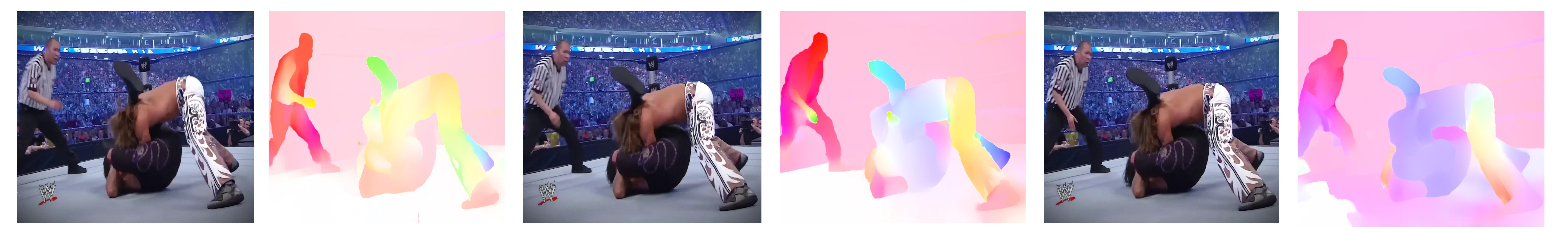}\\[-0.2em]
{\small (a) Scenario 1: intense exercise (real video)}\\[0.6em]
\includegraphics[width=0.9\linewidth]{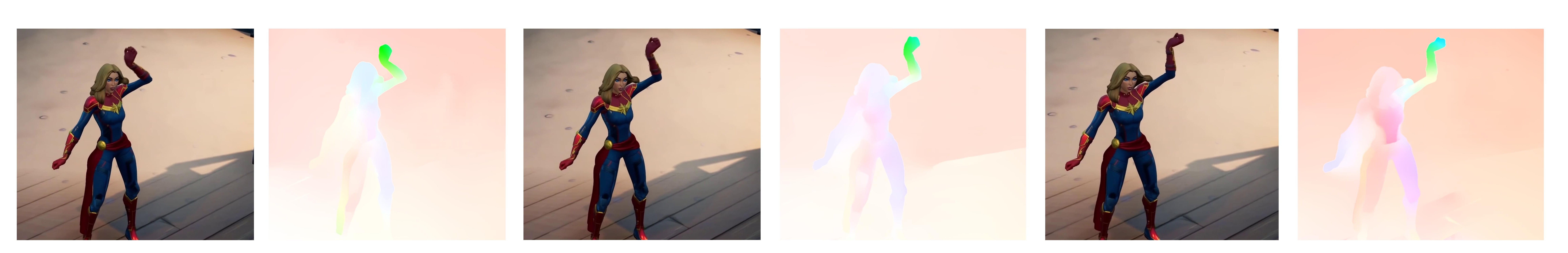}\\[-0.2em]
{\small (b) Scenario 2: character motion (AIGC video)}\\[0.6em]
\includegraphics[width=0.9\linewidth]{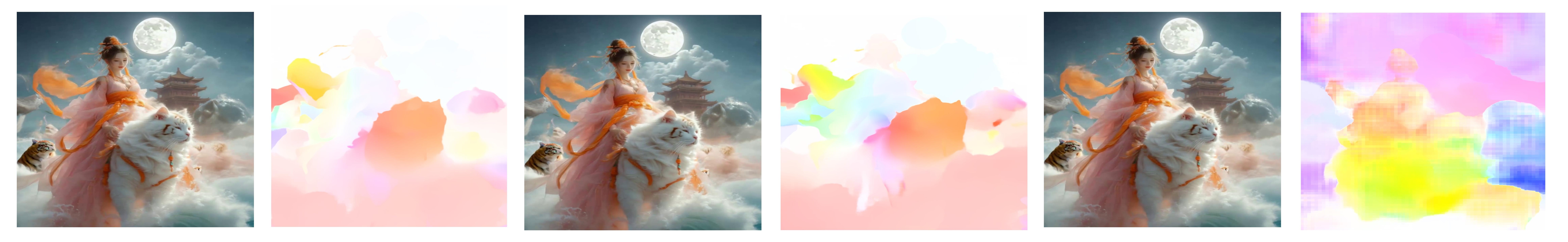}\\[-0.2em]
{\small (c) Scenario 3: fantasy scene (AIGC video)}\\[0.6em]
\includegraphics[width=0.9\linewidth]{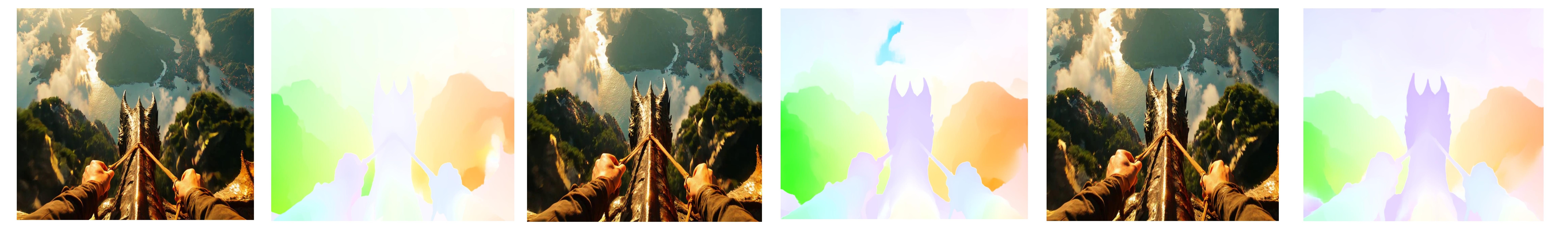}\\[-0.2em]
{\small (d) Scenario 4: first-person view (AIGC video)}
\caption{Optical flow processing results of different video scenarios}
\label{fig:optical_flow_results}
\end{figure}

\begin{enumerate}[label=(\roman*)]
\item The preprocessed video frames, with an input shape of $[B,T,3,H,W]$, are fed into the optical flow branch. These frames are first converted from RGB to grayscale. The dense inverse search (DIS) optical flow algorithm from OpenCV is then used to compute the optical flow fields between consecutive frames. Fig.~\ref{fig:optical_flow_results} visualizes the output of this module on different frame types, with the left side of each subfigure showing the input frame and the right side displaying the computed optical flow field. These fields are generated using the DIS method. In these visualizations, hue represents the direction of motion, while intensity corresponds to its magnitude (speed). Consequently, areas with large motions appear bright and highly saturated, whereas regions with subtle motion are represented by pale colors. This provides a dense, quantifiable representation of motion for feature extraction. The optical flow results directly encode the ``motion information'' within the video. Subsequent modules can then leverage these motion-based features to dynamically assess the likelihood of a video being a forgery.
\end{enumerate}

By contrasting the first row (real video) with the subsequent three rows (AIGC videos) in Fig.~\ref{fig:optical_flow_results}, a fundamental discrepancy in motion modeling becomes evident. While the real scenario exhibits smooth, continuous, and physically grounded temporal transitions, the generated counterparts manifest various motion artifacts. In the AIGC video scenarios, the optical flow fields often show fragmented motion vectors at object boundaries or abrupt changes in intensity that violate the laws of inertia. These ``motion realism clues'' provide a potent diagnostic signal that complements spatial-frequency features, enabling the model to effectively identify subtle temporal inconsistencies.

\begin{enumerate}[label=(\roman*),resume]
\item Motion feature encoding: the normalized optical flow fields are processed by a CNN for feature extraction. This network employs a ``4-Conv + Pooling + Fully Connected'' structure. All four convolutional layers use a $5\times5$ kernel with a stride of 2 and padding of 2, each followed by batch normalization (BN) and a ReLU activation function. An adaptive average pooling layer is then applied to compress the spatial dimensions. The resulting motion features are finally output through a fully connected layer.
\item Temporal modelling of features: the encoded optical flow features are fed into a BiLSTM network for temporal modeling. This module takes the input features in the shape $[B,T-1,2d_f]$. A multi-layer BiLSTM network (default: two layers) is applied to model the inter-frame dependencies of motion, capturing the relationships between motion features across frames. It outputs temporally encoded features with dimensions $[B,T-1,2d_h]$, where $d_h=256$. This stage preserves the complete temporal dynamics of the optical flow sequence, providing discriminative representations for detecting motion anomalies, such as violations of physical laws or abrupt changes between consecutive frames.
\end{enumerate}

\subsection{Cross-modal inconsistency mining via cross-attention}
This module employs a cross-attention mechanism to model the relationship between spatial-frequency and optical flow features. It achieves this by learning an informed weighting scheme that highlights the most salient and complementary components from each modality. The resulting adaptively weighted features provide a robust representation for accurate video classification. The specific implementation of this method is as follows:

\subsubsection{Intuition of asymmetric diagnosis}
We assign asymmetric roles to the two feature streams to reflect a physical cause-and-effect relationship: appearance governs plausible motion.
\begin{enumerate}[label=(\roman*)]
\item Query: ``appearance context''. The spatial-frequency feature $F_{\mathrm{sf}}$ serves as the query. This representation encapsulates the static context of the video frame, effectively posing the question: ``given this object's texture, shape, and frequency signature, what is its probable motion?''
\item Key/Value: ``motion evidence''. The optical flow feature $F_{\mathrm{of}}$ serves as the key and value. This provides the directly observed dynamic evidence, constituting the answer: ``this is the actual motion trajectory we observe.''
\end{enumerate}

\subsubsection{Diagnostic process}
The cross-attention weights $A_{\mathrm{of}\rightarrow\mathrm{sf}}$ directly quantify the compatibility between the motion expected from the appearance context and the motion actually observed in the optical flow. A low attention score in specific regions indicates a critical inconsistency: the observed motion constitutes an implausible response to the appearance context. Thus, the weighted output $F_{\mathrm{of}}^{\mathrm{att}}$ represents ``appearance-rectified motion features'' that are highly discriminative for detecting generated content.
\begin{enumerate}[label=(\roman*)]
\item The inputs into this module are two key sets of features. One is the temporal features from the fusion of spatial features and frequency-domain features that focus on static discriminative information such as texture and structure in the video frames. The other is the output temporal features from the optical flow branch, which characterizes the dynamic patterns of inter-frame motion. Since their output dimensions are different (the spatial-frequency feature has temporal length $T$, while the optical-flow feature has temporal length $T-1$), a linear projection layer is required to transform both inputs so that they match a single unified dimensionality.
\item The spatial-frequency feature acts as the query, while the optical-flow output features act as the key and value. Thus, the mechanism captures supplementary information with regard to motion anomalies that are relevant to the spatial features. In order to preserve the feature information, an elementwise addition is made between the originally projected spatial-frequency features and the attention output. The final output is then normalized to produce the enhanced spatiotemporal features.
\end{enumerate}

\subsubsection{Advantage over baselines}
This method adds a dynamic, interpretable version of relational reasoning. Unlike concatenation or additive fusion, it does not assume that the modalities are equally reliable or directly align individually. It is specifically designed to highlight and use their points of disagreement, which are the Achilles' heel of generative models.

\section{Experimental design and result analysis}
\subsection{Experimental purpose}
The objective of our experiments is to evaluate the effectiveness of the proposed methodology---which integrates spatial, frequency, and optical flow features---in distinguishing real videos from AIGC. A further aim is to demonstrate its performance improvements over existing methods. The model was trained and evaluated using a combination of a private self-built dataset and several public datasets. Performance is assessed through the standard metrics of accuracy, precision, recall, and F1-score. Finally, the model is validated on open-source datasets and compared against existing baseline models.

\subsection{Dataset construction}
For model training, we employed a mixed dataset comprising videos from a proprietary collection and augmented with publicly available sources. This approach was necessary due to our specific hardware constraints, as relying solely on public datasets led to suboptimal results. The final dataset (see Table~\ref{tab:dataset_composition}) contains 1\,312 real and 1\,312 fake videos for training, 448 of each for validation, and 640 real and 703 fake videos for testing. Specifically, the ``fake'' subset integrates open-source samples from GenVidBench (accounting for approximately 71\% of training fakes, including CogVideo and Mora) and official demonstration videos from Sora, Kling, and Hunyuan. The ``real'' subset comprises videos from the HAIC dataset (approximately 38\%) and high-quality clips collected from Mixkit (approximately 62\%). This hybrid composition ensures the model is exposed to both standardized benchmarks and diverse real-world scenarios. For scenarios with limited annotated data---such as scarce AIGC videos from emerging models like Sora---a multi-stage semi-supervised active learning framework has proven effective for improving model robustness by iteratively selecting informative samples for annotation \cite{rf21}. This provides a valuable strategy for expanding our dataset in future work.

\begin{table}[H]
\centering
\caption{Composition of dataset videos}
\label{tab:dataset_composition}
\begin{tabular}{lrrr}
\toprule
Dataset & Real video & Fake video & Total\\
\midrule
Training set & 1\,312 & 1\,312 & 2\,624\\
Validation set & 448 & 448 & 896\\
Test set & 640 & 703 & 1\,343\\
\bottomrule
\end{tabular}
\end{table}

A computationally efficient keyframe selection technique is employed to accelerate training. In this procedure, a unified score for each frame is computed based on its frequency-domain saliency (i.e., mean high-frequency energy) and motion entropy (derived from optical flow vectors). The frames are then ranked by this score, and the top 10 are selected as keyframes. To optimize processing efficiency while retaining temporal density, a sampling interval of two frames is applied during the candidate assessment. To incorporate temporal context, a 7-frame snippet---comprising the keyframe, the three preceding frames, and the three subsequent frames---is extracted for each selected keyframe. For videos shorter than 100 frames, the entire video is used. The resulting unique set of frames, labeled as real or AIGC, constitutes the final dataset for model training.

\subsection{Experimental setup}
The model is implemented in PyTorch and trained on an NVIDIA GPU (with 32 GB of video random access memory (VRAM)) for 24 epochs with a batch size of 16, using cross-entropy loss. The learning rate is warmed up from $1.1\times10^{-6}$ to $1.6\times10^{-5}$ over the first 10 epochs, coupled with an early stopping strategy (patience = 10). The input is featurized by several encoders: a frozen 7-layer ResNet50 (outputting 2\,048-D features), a frequency encoder (256-D), and an optical flow encoder (512-D). Both BiLSTMs are 2-layer models with 256 hidden units each. The cross-attention module uses four heads, has a latent dimension of 512, a dropout rate of 0.3, and produces a 512-dimensional output.

Given the limited scale of our dataset, we employ several data augmentation techniques, including color jittering, random grayscale conversion, Gaussian noise addition, and various flipping operations. The hybrid architecture of our model---which integrates CNNs for spatial and optical-flow feature extraction, LSTMs for temporal modeling, and attention mechanisms for feature enhancement---has proven effective for diverse spatiotemporal prediction tasks. These applications include carbon emission measurement \cite{rf22}, heat load prediction \cite{rf23}, IoT attack detection \cite{rf24}, cryptocurrency forecasting \cite{rf25}, chaotic time series analysis \cite{rf26}, resilience recovery forecasting in complex traffic networks \cite{rf27}, and active defense mechanisms for system health assessment in safety-critical domains \cite{rf28,rf29,rf30}. The model's success across these diverse domains demonstrates the rationality of its architectural design and underscores its strong generalization capability.

\subsection{Experiment design comparison}
To comprehensively evaluate the effectiveness of our proposed multimodal fusion model for AIGC video authentication, we conducted a series of comparative experiments. All experiments were conducted under identical software and hardware environments, using the same hyperparameter settings, datasets, and evaluation metrics. To ensure reliable comparisons, only one variable was altered at a time across the experiments. The experimental design is described in detail below.

\subsubsection{Comparison of different data loading methods}
In this study, the data loading method plays a critical role in the model training pipeline. Its design directly influences both the model's efficiency in extracting meaningful information from videos and the associated computational cost. Unlike image data, videos contain rich temporal information. An inefficient loading method can result in the loss of temporal features and increased training time due to redundant data processing. This experiment compares the ``pre-screened keyframes'' and ``complete video input'' loading methods. While maintaining spatial feature optimization and controlling all other variables, we focus exclusively on the data loading strategy. Evaluation metrics include model accuracy, training duration, and memory usage to comprehensively assess the advantages of each approach.

\subsubsection{Comparative experiment}
To evaluate the effectiveness of the proposed CrossAtt-VFD framework, we conduct a comprehensive performance comparison against several representative SOTA models. All models are trained and tested under identical hardware environments and hyperparameter settings to ensure fairness. The following baselines are selected.
\begin{enumerate}[label=(\roman*)]
\item CNN-LSTM-Attention: a classic spatiotemporal architecture that uses a CNN backbone for frame feature extraction, followed by an attention-enhanced LSTM to model temporal dependencies. It represents the standard approach for video forgery detection without multi-modal fusion.
\item EfficientFormer: a vision transformer architecture optimized for high-speed performance, primarily focusing on extracting complex spatial artifacts.
\item ViTranSP: a specialized video transformer designed to capture long-range spatiotemporal relationships by processing video patches \cite{rf35}.
\end{enumerate}

\subsubsection{Ablation experiment comparison}
The three modalities---spatial, frequency domain, and optical flow---each capture distinct features in videos. This experiment establishes the performance ceiling of each modality by training separate baseline models, providing a benchmark for the subsequent fusion model. All modalities use identical classification heads and training parameters. Thus, any performance differences can be attributed solely to the modalities themselves.

\subsection{Experimental results and analysis}
\subsubsection{Results of a comparative experiment on different data loading methods}
This study compares two data loading strategies: ``complete video input'' and ``pre-screened keyframes'', under identical experimental conditions. Repeated trials produced the following results.

\begin{table}[H]
\centering
\caption{Comparison of training metrics for two different data loading methods}
\label{tab:data_loading_comparison}
\begin{tabular}{lcc}
\toprule
Index & Selecting keyframes & Entering full video directly\\
\midrule
Loss function & $0.2115\pm0.0024$ & $0.2244\pm0.0030$\\
Accuracy rate & $93.30\%\pm0.17\%$ & $92.50\%\pm0.26\%$\\
Precision rate & $91.28\%\pm0.71\%$ & $90.73\%\pm0.66\%$\\
Recall rate & $94.25\%\pm0.58\%$ & $93.00\%\pm0.71\%$\\
F1-score & $92.74\%\pm0.45\%$ & $91.85\%\pm0.32\%$\\
VRAM usage & $9539\mathrm{MiB}\pm17\mathrm{MiB}$ & $9693\mathrm{MiB}\pm26\mathrm{MiB}$\\
Time per training round & $4\ \mathrm{min}\ 25\ \mathrm{s}\pm12\ \mathrm{s}$ & $6\ \mathrm{min}\ 17\ \mathrm{s}\pm13\ \mathrm{s}$\\
\bottomrule
\end{tabular}
\end{table}

\begin{figure}[H]
\centering
\includegraphics[width=\linewidth]{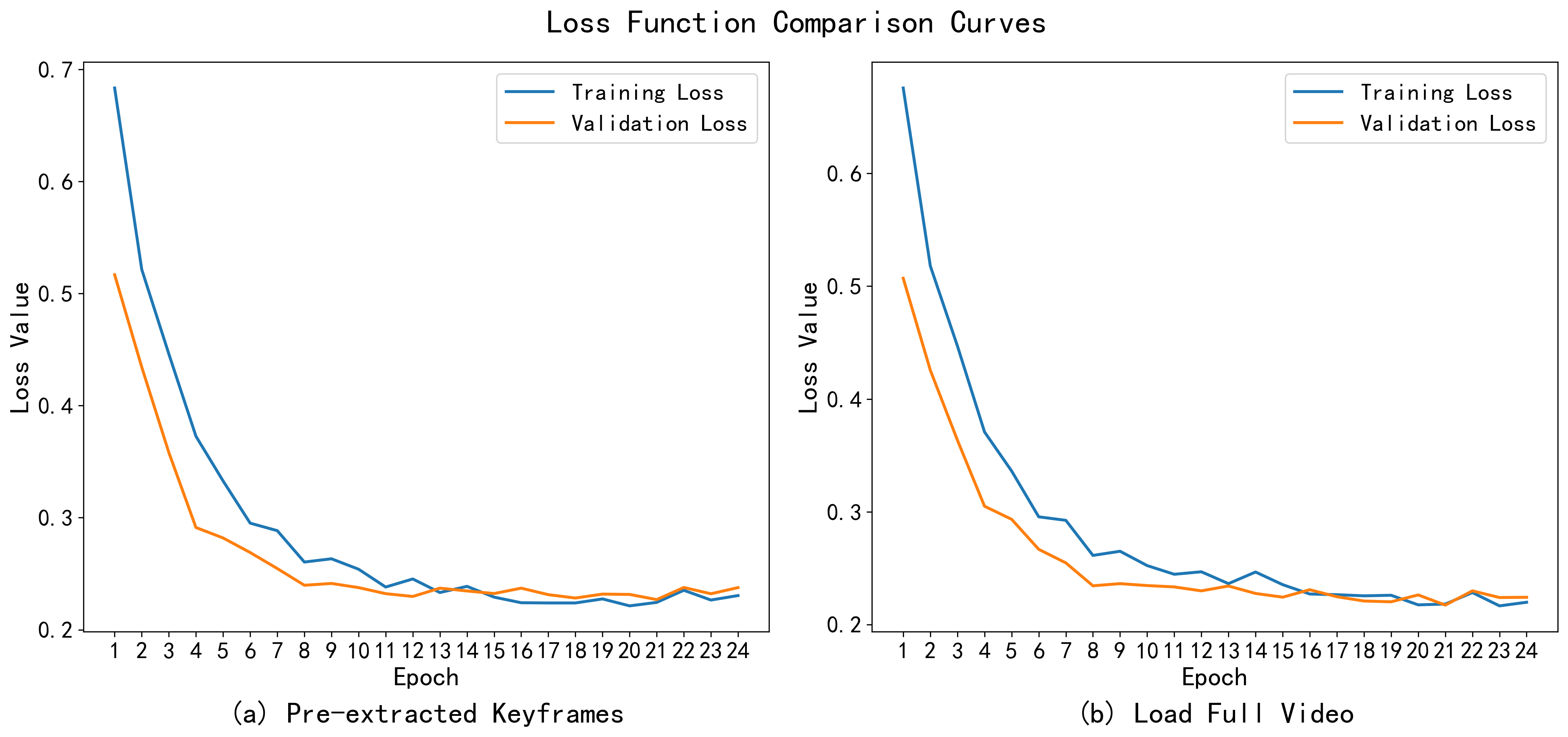}
\caption{Training loss curves for different loading methods}
\label{fig:loading_loss}
\end{figure}

As summarized in Table~\ref{tab:data_loading_comparison} and illustrated by the training loss curves in Fig.~\ref{fig:loading_loss}, both methods converge successfully without showing signs of overfitting, demonstrating their ability to learn discriminative video features. However, the ``pre-screened keyframes'' approach consistently outperforms the ``complete video input'' baseline across all core metrics: $+0.8\%$ in accuracy, $+0.55\%$ in precision, $+1.25\%$ in recall, and $+0.89\%$ in F1-score. Furthermore, it reduces the training time per epoch by 112 s, demonstrating significantly improved computational efficiency. This performance advantage can be attributed to the keyframe selection mechanism, which filters out redundant or static frames. Training on consecutive frames from complete videos may distract the model with repetitive scene content or minor, non-discriminative variations. In contrast, the keyframe strategy directs the model's attention to the most informative temporal segments---those most likely to contain the subtle artifacts characteristic of AIGC. Consequently, we adopt the ``pre-screened keyframes'' method for all subsequent experiments.

\subsubsection{Comparative experiment results}
\begin{figure}[H]
\centering
\includegraphics[width=\linewidth]{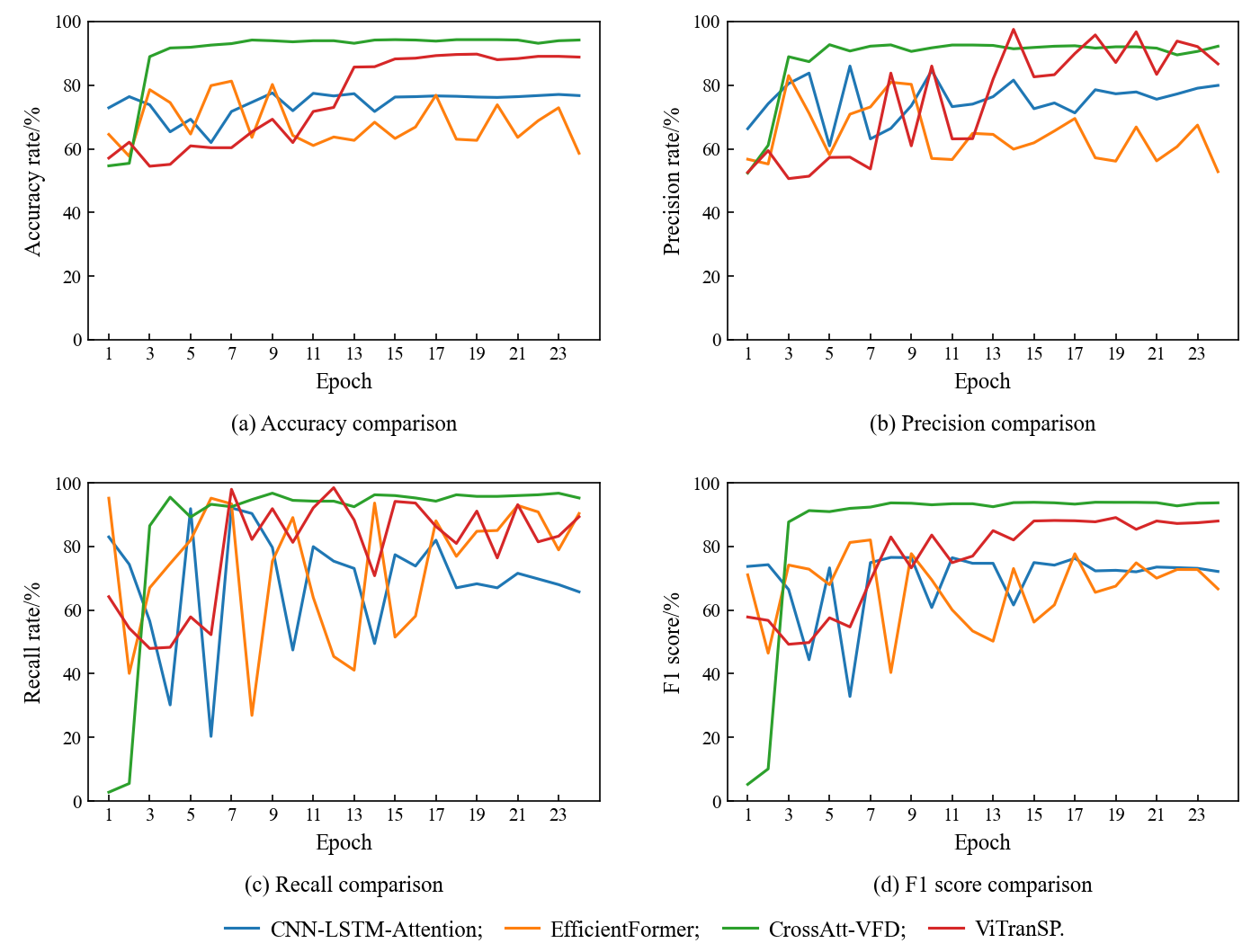}
\caption{Performance metric comparison curves of different detection models}
\label{fig:sota_curves}
\end{figure}

The quantitative results recorded at each model's best validation epoch are summarized in Table~\ref{tab:sota_comparison}. Furthermore, Fig.~\ref{fig:sota_curves} visualizes the performance fluctuations and convergence curves, providing a comprehensive view of the training dynamics for each baseline.

\begin{table}[H]
\centering
\caption{Quantitative comparison with SOTA methods}
\label{tab:sota_comparison}
\begin{tabular}{lcccc}
\toprule
Model & Accuracy (\%) & Precision (\%) & Recall (\%) & F1-score (\%)\\
\midrule
CNN-LSTM-Attention & 77.56 & 73.54 & 79.70 & 76.50\\
EfficientFormer & 81.28 & 73.16 & 93.40 & 82.05\\
ViTranSP & 89.77 & 87.14 & 91.12 & 89.09\\
\textbf{CrossAtt-VFD} & \textbf{94.32} & \textbf{91.67} & \textbf{96.25} & \textbf{93.90}\\
\bottomrule
\end{tabular}
\end{table}

The quantitative results presented in Table~\ref{tab:sota_comparison} and the training dynamics illustrated in Fig.~\ref{fig:sota_curves} demonstrate the definitive advantages of the proposed CrossAtt-VFD framework over representative SOTA baselines. CrossAtt-VFD achieves a validation accuracy of 94.32\%, marking a 4.55\% improvement over the strongest baseline, ViTranSP (89.77\%). A particularly significant finding is the breakthrough in precision, where our model reaches 91.67\%. In comparison, CNN-LSTM-Attention and EfficientFormer exhibit much lower precision (73.54\% and 73.16\%, respectively), suggesting that these models frequently misclassify authentic videos as forgeries when encountering high-fidelity textures. This indicates that relying solely on spatial artifacts or simple temporal modeling is insufficient for distinguishing sophisticated AIGC. CrossAtt-VFD effectively addresses this false-positive challenge by cross-verifying appearance features with motion dynamics. As shown in the performance curves, traditional spatiotemporal models like CNN-LSTM-Attention and EfficientFormer display severe fluctuations during the training process. These fluctuations reflect an instability in capturing the subtle, evolving footprints of synthetic videos across different epochs. Conversely, CrossAtt-VFD exhibits superior convergence stability and robustness. This stability stems from the cross-attention mechanism, which does not merely concatenate features but acts as an ``inconsistency diagnostic engine''. The performance gap between ViTranSP and CrossAtt-VFD highlights that global spatiotemporal attention alone cannot fully resolve the nuanced discrepancies inherent in generative models. The success of CrossAtt-VFD validates our core hypothesis: cross-modal inconsistency (e.g., motion that is statistically or physically inconsistent with the visual scene) provides a more potent and reliable diagnostic signal than artifacts found within any single modality. This allows the framework to maintain high accuracy even when the generative model successfully replicates realistic textures in the spatial domain.

\subsubsection{Ablation experiment comparison results}
\begin{figure}[H]
\centering
\includegraphics[width=\linewidth]{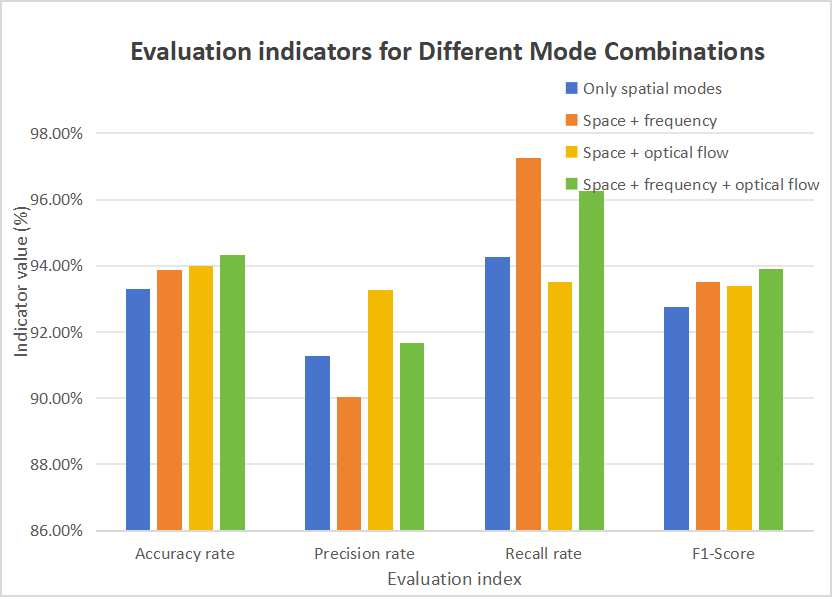}
\caption{Evaluation indicators for different mode combinations}
\label{fig:ablation_metrics}
\end{figure}

To quantitatively assess the contribution of each modality, we conducted a comprehensive ablation study with four experimental settings: spatial-only, spatial + frequency, spatial + optical flow, and the full triple-modality fusion. All configurations shared identical hyperparameters and classification heads, ensuring that performance differences were solely attributable to the input modalities. The results are presented in Table~\ref{tab:ablation_comparison} and Fig.~\ref{fig:ablation_metrics}, with the latter visualizing the comparative trends of accuracy, precision, recall, and F1-score across different modal combinations.

\begin{table}[H]
\centering
\caption{Comparison of training indicators for different modal combinations}
\label{tab:ablation_comparison}
\resizebox{\linewidth}{!}{%
\begin{tabular}{lcccc}
\toprule
Indicator & Only spatial mode & Spatial + frequency domain & Spatial + optical flow & Full triple-modality\\
\midrule
Loss function & $0.2115\pm0.0024$ & $0.2027\pm0.0033$ & $0.1910\pm0.0022$ & $0.2252\pm0.0026$\\
Accuracy rate & $93.30\%\pm0.17\%$ & $93.86\%\pm0.24\%$ & $93.98\%\pm0.14\%$ & $94.32\%\pm0.10\%$\\
Precision rate & $91.28\%\pm0.71\%$ & $90.05\%\pm0.99\%$ & $93.27\%\pm0.56\%$ & $91.67\%\pm0.45\%$\\
Recall rate & $94.25\%\pm0.58\%$ & $97.25\%\pm1.39\%$ & $93.50\%\pm0.40\%$ & $96.25\%\pm0.57\%$\\
F1-score & $92.74\%\pm0.45\%$ & $93.51\%\pm0.28\%$ & $93.38\%\pm0.17\%$ & $93.90\%\pm0.10\%$\\
VRAM usage & $9539\mathrm{MiB}\pm17\mathrm{MiB}$ & $9693\mathrm{MiB}\pm12\mathrm{MiB}$ & $13830\mathrm{MiB}\pm10\mathrm{MiB}$ & $13978\mathrm{MiB}\pm8\mathrm{MiB}$\\
Time per training round & $4\ \mathrm{min}\ 25\ \mathrm{s}\pm12\ \mathrm{s}$ & $5\ \mathrm{min}\ 7\ \mathrm{s}\pm18\ \mathrm{s}$ & $5\ \mathrm{min}\ 30\ \mathrm{s}\pm12\ \mathrm{s}$ & $5\ \mathrm{min}\ 48\ \mathrm{s}\pm14\ \mathrm{s}$\\
\bottomrule
\end{tabular}}
\end{table}

\begin{enumerate}[label=(\roman*)]
\item Spatial modality only: stable baseline performance with identification gaps. This baseline configuration established a balanced performance, achieving 93.30\% accuracy and a 92.74\% F1-score. However, its relatively lower precision reveals a limitation: reliance on static appearance alone can lead to false positives, misclassifying authentic videos with unusual textures as fake.
\item Spatial + frequency domain: enhanced recall at the cost of precision imbalance. Integrating frequency-domain analysis pushed the model's recall to 97.25\%, indicating a heightened sensitivity to detecting AIGC videos. However, this came at the cost of reduced precision, suggesting that some real videos with frequency-domain characteristics similar to generated content were incorrectly flagged, creating an accuracy-recall trade-off.
\item Spatial + optical flow: marked precision improvement with slight recall dip. The inclusion of optical flow features had a complementary effect, boosting precision to 93.27\%. This demonstrates that motion cues are highly effective for identifying physical implausibilities, thereby reducing false alarms. A slight dip in recall implies that some AIGC videos with convincing motion but flawed textures may be missed. This configuration also achieved the lowest training loss, indicating highly stable feature learning.
\item Spatial + frequency domain + optical flow: optimal comprehensive performance with strong metric balance. The complete model, leveraging our cross-attention-based fusion, achieved the most robust and balanced performance. Its high accuracy (94.32\%) and F1-score (93.90\%) underscore the principle that the most powerful detection signals often emerge from the inconsistencies between modalities. The cross-attention mechanism acts as a diagnostic engine, identifying instances where, for example, a subtle texture anomaly (captured by frequency) coincides with an implausible motion (captured by optical flow) for the same object. This synergistic $1+1+1>3$ effect effectively closes the detection blind spots inherent in single-modality or naive fusion approaches.
\end{enumerate}

In conclusion, the ablation experiments confirm the core logic of feature enhancement: efficacy stems from modal complementarity rather than mere quantity. Specifically, the frequency domain contributes spectral anomaly clues that significantly improve recall, while the optical-flow modality provides motion realism clues that optimize precision. The full tri-modal combination achieves a synergistic effect ($1+1+1>3$), covering the blind spots of single- and dual-modal approaches and realizing a balanced optimization of performance.

\subsection{Model's effect display in the test set}
\begin{figure}[H]
\centering
\includegraphics[width=0.78\linewidth]{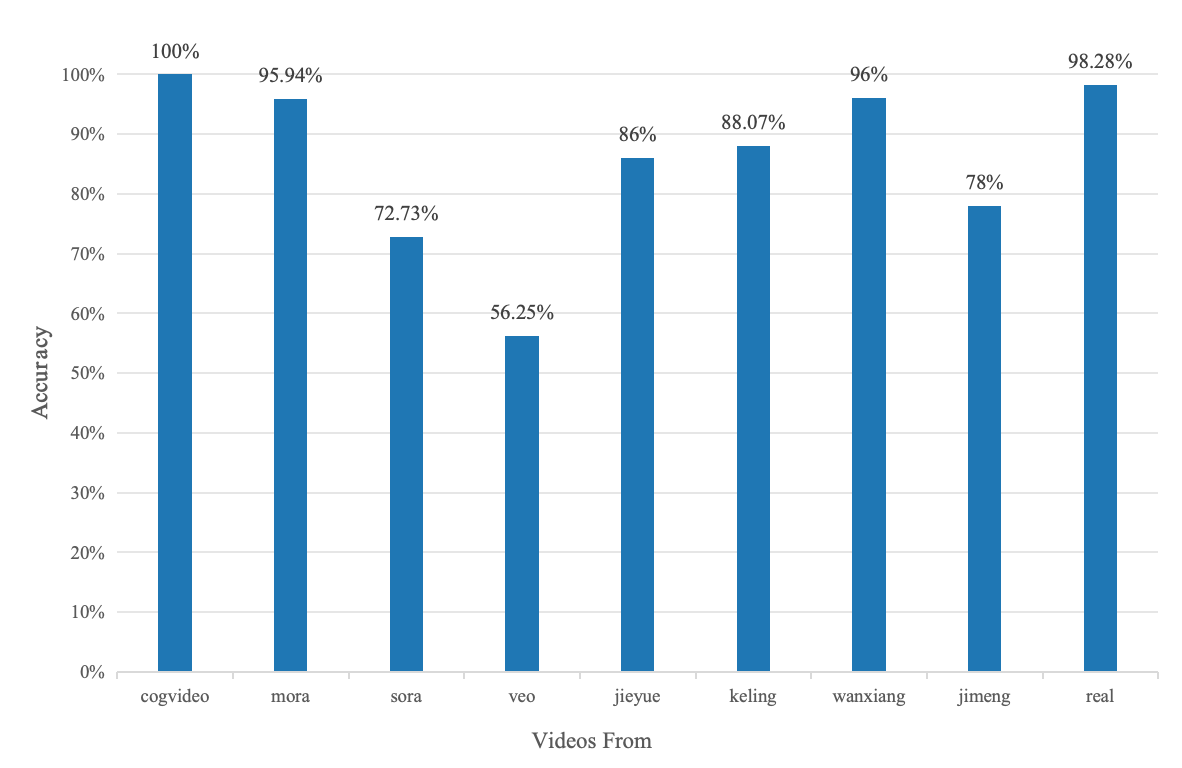}
\caption{Display of test set detection results}
\label{fig:test_set_results}
\end{figure}

The final tri-modal model was rigorously evaluated on the held-out test set. As shown in Fig.~\ref{fig:test_set_results}, the model demonstrated strong detection performance for videos generated by models included in the training data (e.g., CogVideo and Hunyuan). It also achieved respectable accuracy (86\%) on videos generated by ``Jieyue'', a source not seen during training. However, its performance on the latest generative video models (Sora, Veo, Jimeng) indicates that significant improvement is needed in detection accuracy---an expected challenge given the rapid pace of advancement in generative technology.

\begin{figure}[H]
\centering
\includegraphics[width=0.78\linewidth]{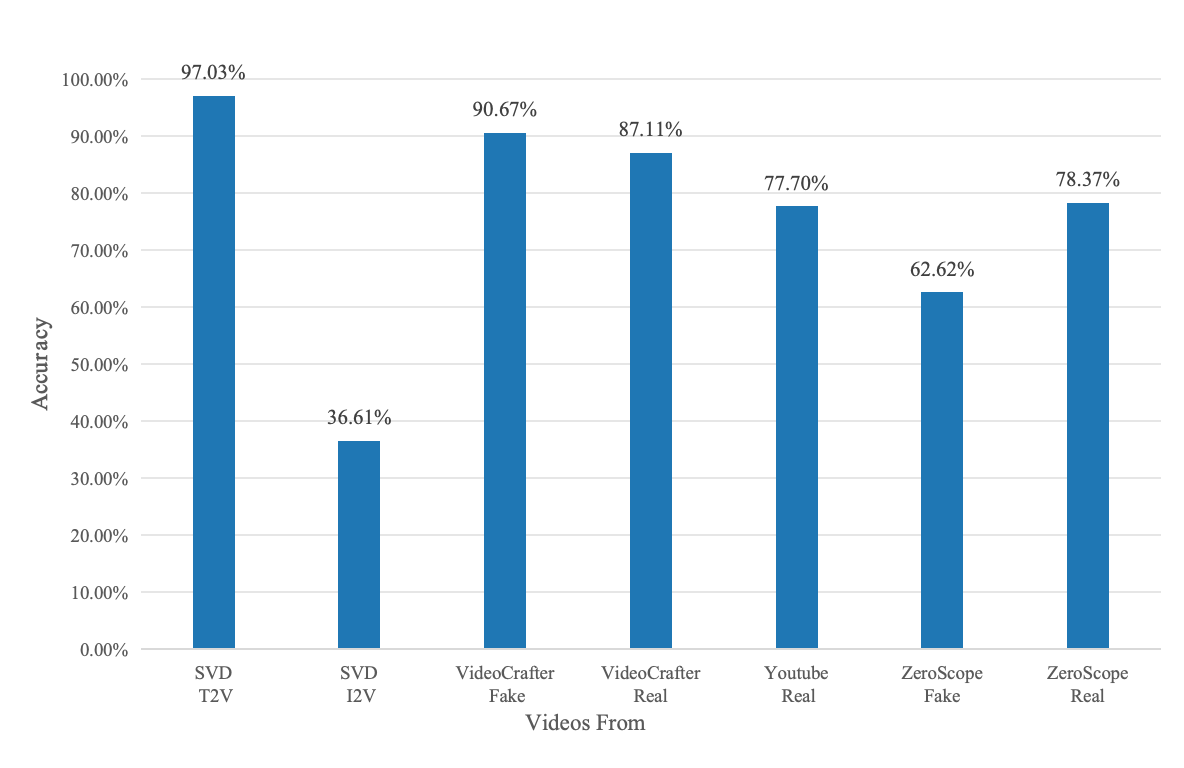}
\caption{Detection results on the open-source dataset DVF}
\label{fig:dvf_results}
\end{figure}

Additionally, the model is tested on the open-source DVF dataset without any fine-tuning, as illustrated in Fig.~\ref{fig:dvf_results}. The results reflect good generalization capability, with an accuracy of 97.03\% on the ``Stable video'' category and 90.67\% on videos generated by VideoCrafter1.

A notable exception is the stable video diffusion image-to-video category, for which the accuracy is low (36.61\%). This is because of the distributional shift caused by this generation paradigm, which is not included in our training dataset. This finding highlights a current limitation and, more importantly, clarifies a future direction for work: to provide the model with data from a wider variety of generation empirical distributions, with a view to improving the robustness of the model, including advanced versions such as VideoCrafter2 \cite{rf37}. The performance profile confirms that, although the current detector affords high efficiency against a variety of AIGC videos, the generalization capability still needs to be continuously broadened. This would keep pace with better generative models as they emerge.

\section{Summary}
This study introduces an efficient framework for detecting AIGC videos by identifying cross-modal inconsistencies. Recognizing that artifacts in synthetic videos often arise from discrepancies between static appearance and dynamic motion, we develop a cross-attention mechanism. This module explicitly highlights these discrepancies by correlating distinctive spatial-frequency features with optical-flow patterns, thereby pinpointing violations of physical motion constraints. Our approach demonstrates diagnostic capabilities that are markedly superior to simple feature-fusion techniques used in existing models.

The framework's effectiveness is validated on a comprehensive dataset comprising diverse SOTA generative models. To improve computational efficiency, we implemented a salient, motion-entropy-based keyframe selection strategy. Ablation studies confirm that while each modality (spatial, frequency, and optical flow) contributes significantly, their enhancement through our cross-attention mechanism yields substantial gains. The full model achieves an accuracy of 94.32\% on our benchmark dataset, demonstrating a significant improvement over obsolete single-modality or simple-fusion baselines by effectively addressing their critical limitations.

Future work will focus on incorporating data from emerging generative models to enhance generalization. Specifically, considering the rapid iteration of generative models and the scarcity of samples for the latest generators (e.g., Sora), we plan to explore few-shot learning or semi-supervised active learning strategies. This would allow the model to adapt quickly to new generation paradigms with limited annotated data, thereby improving its extendability in small-sample scenarios. Furthermore, the cross-attention mechanism itself offers opportunities for refinement and integration into real-time authentication systems, promising broader practical application.

\bibliography{references}

\clearpage
\section*{Biographies}

\bioentry{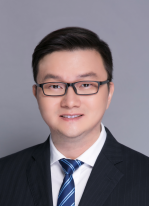}{HONG Sheng}%
{was born in 1981. He received his Ph.D. degree from Beihang University, China, in 2009. He is currently an associate professor and doctoral supervisor at Beihang University. His research interests include artificial intelligence and big data, artificial intelligence-driven cyber security and industrial internet.}%
{shenghong@buaa.edu.cn}

\bioentry{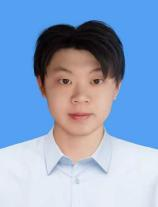}{WANG Xuanqi}%
{was born in 2002. He received his bachelor's degree in engineering from Inner Mongolia University. Currently, he is pursuing his master's degree at the School of Information Engineering of Nanchang University. His research interests include generative artificial intelligence, video detection, and multimodal feature extraction.}%
{15894886025@163.com}

\bioentry{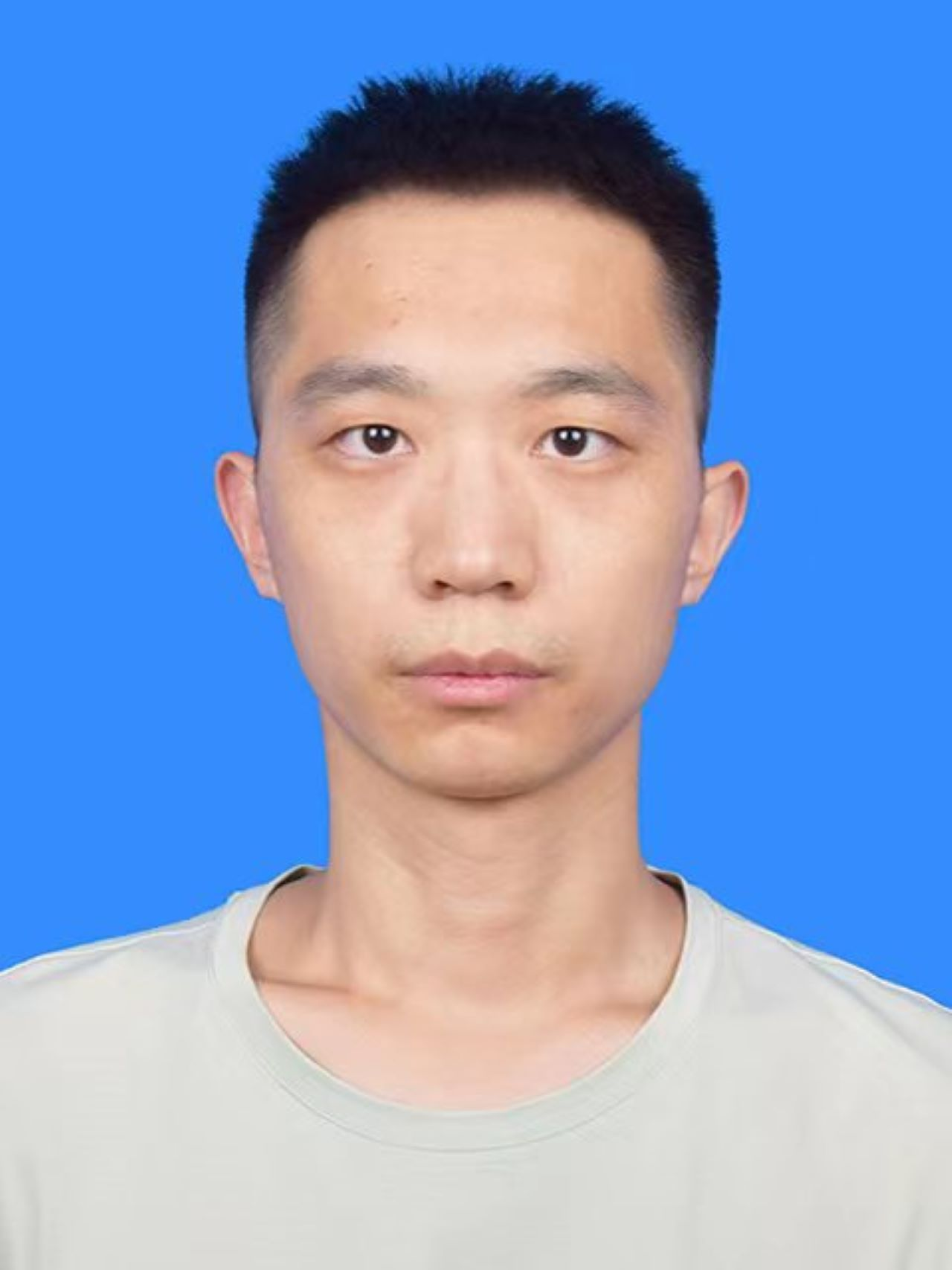}{ZHANG Chang}%
{was born in 1990. He received his bachelor's degree in engineering from Beihang University in Beijing in 2025. His research interests include optimization of machine learning models and cross-modal data fusion.}%
{1098033100@qq.com}

\bioentry{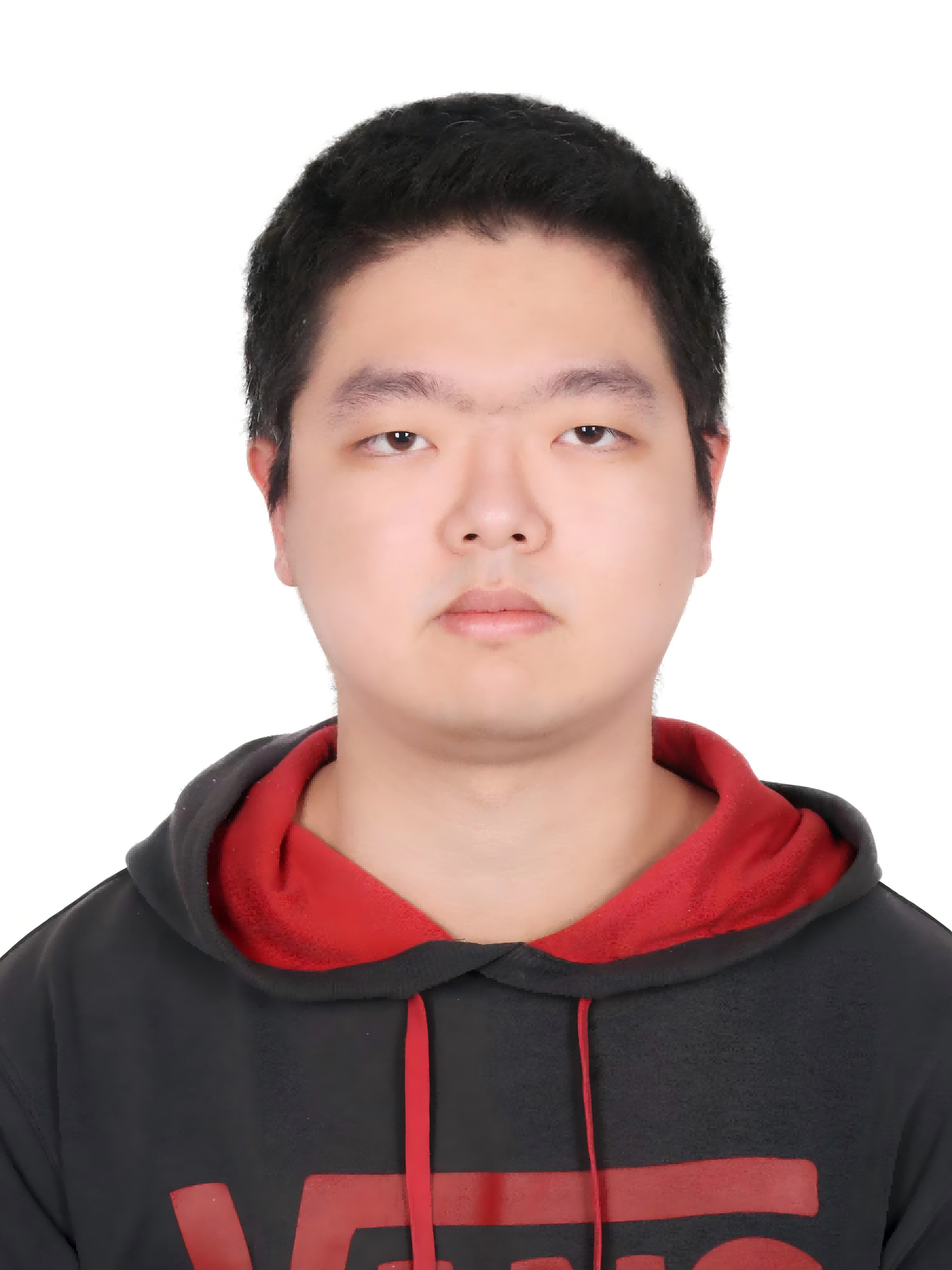}{WANG Jiacheng}%
{was born in 2001. He received his B.S. degree in information security from Lanzhou University, China, in 2023. He is pursuing his M.Eng. degree with the School of Cyber Science and Technology, Beihang University, China. His research interests include artificial intelligence, information security, and applied cryptography.}%
{wjc1321@163.com}

\bioentry{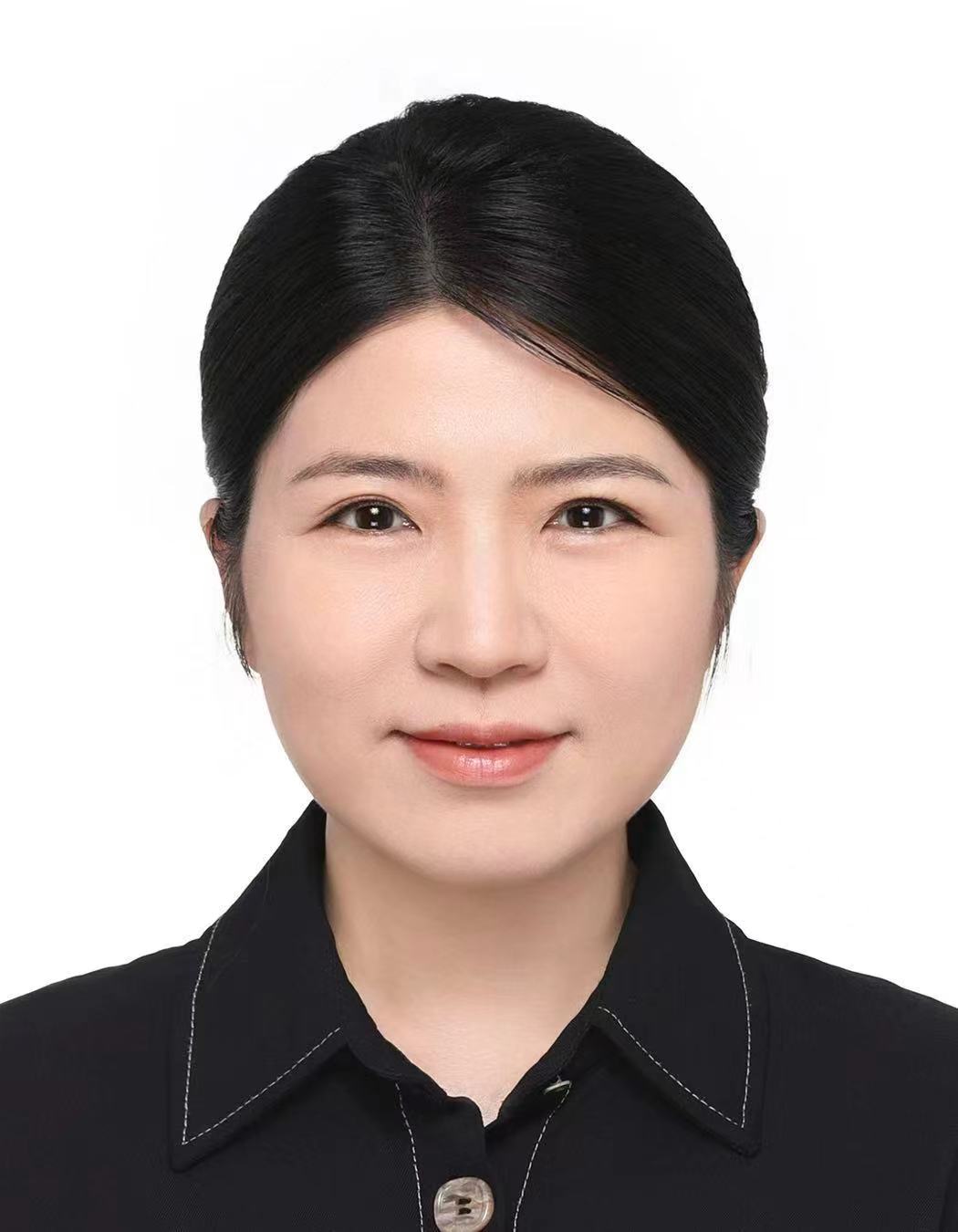}{DUAN Pingxia}%
{was born in 1980. She received her bachelor's degree in management information systems from Chongqing University of Technology in 2001, and her M.S. degree in business administration from Chongqing University in 2004. Her research interest focuses on data security governance and management systems, and video forgery detection.}%
{409821331@qq.com}

\bioentry{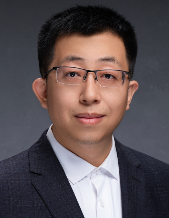}{WANG Yuwei}%
{was born in 1980. He received his Ph.D. degree from the University of Chinese Academy of Sciences in 2020, majoring in computer science and technology. His current research interests include edge intelligence and AI-driven cyber security.}%
{ywwang@ict.ac.cn}

\end{document}